\documentclass[10pt,twocolumn,letterpaper]{article}

\usepackage{iccv}
\usepackage{times}
\usepackage{epsfig}
\usepackage{graphicx}
\usepackage{amsmath}
\usepackage{amssymb}

\usepackage{gensymb}
\usepackage{booktabs}
\usepackage{multirow}
\usepackage{xcolor}

\usepackage[pagebackref=true,breaklinks=true,letterpaper=true,colorlinks,bookmarks=false]{hyperref}

\iccvfinalcopy 

\def\iccvPaperID{9202} 

\ificcvfinal\fi

\begin{document}

\title{NPR: Nocturnal Place Recognition in Streets}

\author{Bingxi Liu$^{1,2}$, Yujie Fu$^{3}$, Feng Lu$^{2, 4}$, Jinqiang Cui$^{2}$, Yihong Wu$^{3}$, Hong Zhang$^{1,*}$\\
$^{1}$Southern University of Science and Technology, Shenzhen, China.\\
$^{2}$Peng Cheng Laboratory, Shenzhen, China.\\
$^{3}$Institute of Automation, Chinese Academy of Sciences, Beijing, China.\\
$^{4}$Tsinghua University, Shenzhen, China.\\
{\tt\small liubx@pcl.ac.cn, hzhang@sustech.edu.cn}
}
\maketitle
\ificcvfinal\thispagestyle{empty}\fi

\begin{abstract}
Visual Place Recognition (VPR) is the task of retrieving database images similar to a query photo by comparing it to a large database of known images. In real-world applications, extreme illumination changes caused by query images taken at night pose a significant obstacle that VPR needs to overcome. However, a training set with day-night correspondence for city-scale, street-level VPR does not exist. To address this challenge, we propose a novel pipeline that divides VPR and conquers Nocturnal Place Recognition (NPR). Specifically, we first established a street-level day-night dataset, NightStreet, and used it to train an unpaired image-to-image translation model. Then we used this model to process existing large-scale VPR datasets to generate the VPR-Night datasets and demonstrated how to combine them with two popular VPR pipelines. Finally, we proposed a divide-and-conquer VPR framework and provided explanations at the theoretical, experimental, and application levels. Under our framework, previous methods can significantly improve performance on two public datasets, including the top-ranked method.

\end{abstract}

\section{Introduction}
\label{sec:introduction}
Visual Place Recognition (VPR) is a fundamental task in the fields of computer vision \cite{netvlad, patch_netvlad, 3d_model, 247_dataset, repetitive_structures, vpr_bench} and robotics \cite{kitti, tro_survey, robotcar, seqslam, appsvr, msls}, which involves returning database images that are similar to a query image by comparing it with a known large-scale database image. As previously reported in research, challenges for VPR include database scale \cite{cosplace}, repeated structures \cite{repetitive_structures}, structural changes \cite{change}, occlusion \cite{netvlad}, viewpoint\cite{viewpoint}, visual scale \cite{scalenet}, illumination \cite{todaygan, delg, robotcar, tro_survey, lamar, long_vl, 247_dataset, msls, vpr_bench}, and seasonal changes \cite{seqslam, nordland}. Almost all recent VPR methods are learning on large-scale datasets \cite{ cosplace, st_lucia, GLDv1, 247_dataset, gldv2}. A neural network is often used to map images into an embedding space that can efficiently distinguish images captured from different places. However, existing datasets restrict the progress of VPR in nighttime street scenes.

\begin{figure}[tbp] 
	\center
	{\includegraphics[width=0.47 \textwidth] {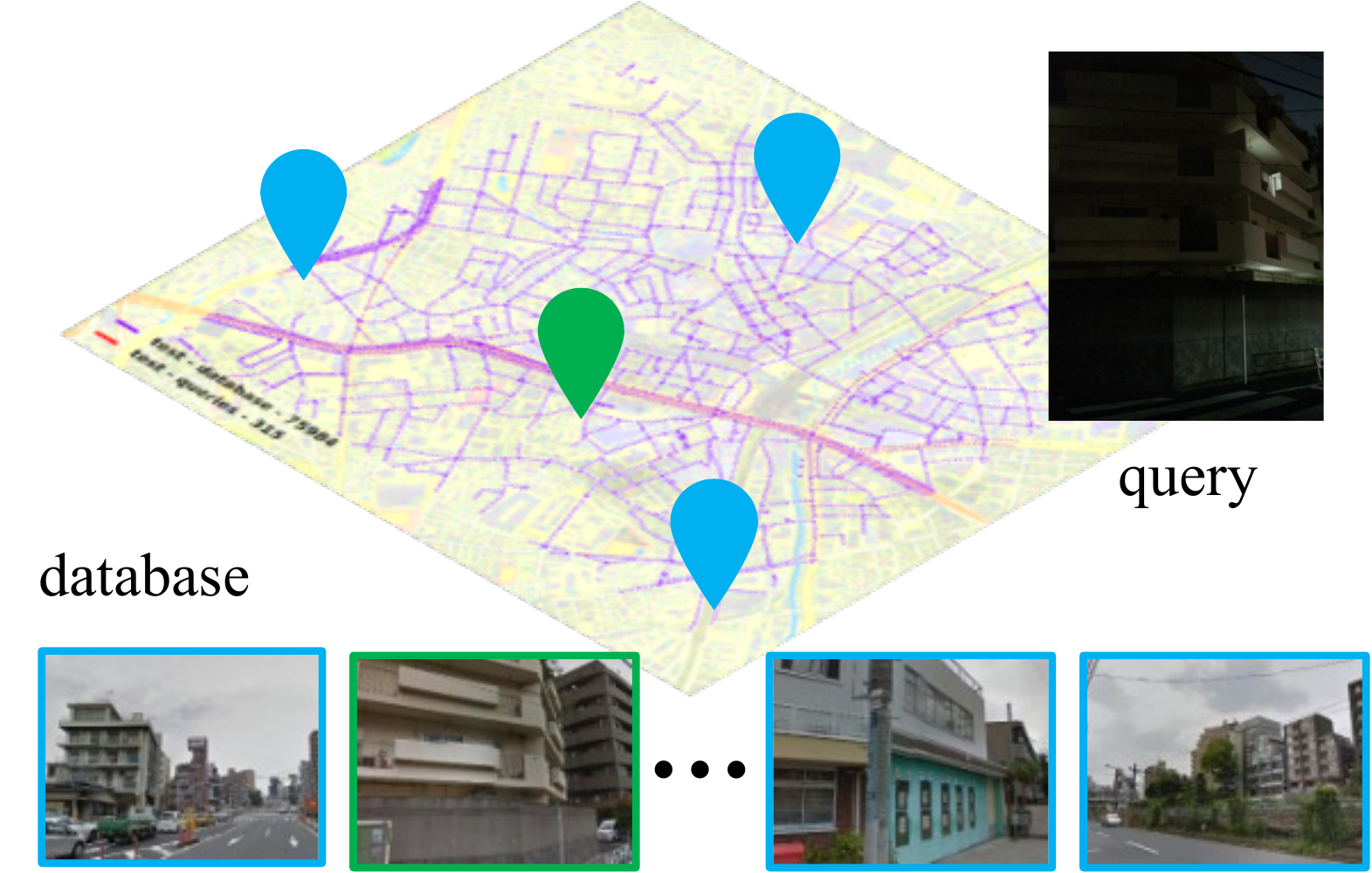}}
	\caption{\textbf{A demo for NPR.} A 1.2km $\times$ 1.2km satellite map is presented, where densely purple dots represent the locations where the database images were captured. Our proposed method can retrieve the correct results from a database of 75,984 images, even when provided with a nighttime captured image that includes a significant change in the view direction.} 
\label{demo1}
\end{figure}

\noindent \textbf{Training sets.} 
Previous research has established large-scale VPR training sets using the downloading interface of Google Street View \cite{gsv_cities, gsv, netvlad,  cosplace}. These training sets were collected using either car-mounted panoramic cameras or pedestrian-carried street-view cameras and covered almost all of the challenges above, except for nighttime scenes. Some VPR training sets for autonomous driving scenes include nighttime scenes but lack other challenges  \cite{robotcar, msls}. For example, cars are limited to roads and mostly use forward-facing cameras, resulting in a lack of view direction changes.

\noindent \textbf{Testing sets.}
 Two representative testing sets for VPR in nighttime street scenes are Tokyo 24/7 \cite{247_dataset} and Aachen Day/Night. i) The Tokyo 24/7 dataset comprises daytime, sunset, and nighttime scenes. However, all research \cite{netvlad, cosplace, dvg, sfrs, sare} has yet to separate these scenes for testing purposes; even the original paper \cite{247_dataset} tested only sunset and nighttime together. This oversight obscures the fact that VPR performance is poor at night. ii) The Aachen Day/Night dataset evaluates Visual Localization (VL) but can also assess VPR as it serves as the first stage in a two-stage VL approach \cite{superglue, hloc, vl_benchmarking}. The dataset is divided into daytime and nighttime and ranked on an evaluation system without visible ground truth. The top-ranked method used VPR to recall 50 candidate frames, which is poor in practical applications.

Under the limitations of the training set, a straightforward idea is to apply Image Enhancement (IE) or image-to-image translation for the nighttime queries to daytime queries. However, these methods or their application in VPR have the following problems: i) These methods introduce additional computing resources and time. ii) Learning-based IE's training sets usually pairs of low-exposure to high-exposure images rather than night-to-day pairs \cite{lol_dataset}, even not outdoors, so they may have poor performance on the VPR dataset.  iii) Inaccurate or erroneous night-to-day image-to-image translation can result in a degradation of VPR performance \cite{todaygan}.

In this article, we successfully address training set issues through reverse "Night-to-Day" and propose a method to divide VPR and conquer Nocturnal Place Recognition (NPR). To summarize, our contributions are as follows:

\begin{itemize}
  \item We propose a dataset comprising street scene images captured during daytime and nighttime and trained an unpaired image-to-image translation network on this dataset.
  \item Using the above translation network, we processed existing VPR datasets to generate the VPR-Night datasets and demonstrated two popular VPR pipelines on how to leverage the VPR-Night datasets.
  \item We propose a divide-and-conquer VPR framework and provide theoretical, experimental, and practical explanations for the framework. Furthermore, under this framework, previous methods exhibit superior performance on public datasets.
\end{itemize}

\section{Related Work}
\label{sec:related_work}

\noindent \textbf{Visual Place Recognition} has been dominated by deep learning methods. Previous research can be summarized in three main aspects: model, loss function, and data. At the model level, Convolutional Neural Networks (CNNs) \cite{netvlad} or Visual Transformers (ViTs) \cite{transvpr} are typically used as the feature extraction network backbone, followed by a pooling or aggregation layer \cite{netvlad, dir}, such as NetVLAD. At the loss function level, triplet loss and contrastive loss are commonly used to increase the Euclidean margin for better feature embedding \cite{dvg}. However, triplet loss has a significant issue with mining hard negative samples. To address this problem, Liu et al. \cite{sare} proposed the statistical attention embedding loss, which efficiently utilizes multiple negative samples for supervision. Moreover, building on \cite{sfrs}, Ge et al. \cite{sfrs} utilized network output scores as self-supervised labels and achieved new state-of-the-art results through iterative training. Berton et al. \cite{cosplace} used the Large Margin Cosine Loss (LMCL) in VPR tasks and demonstrated superior performance over triplet loss. At the data level, it can be further divided into two scenarios: road scenes and street scenes. Road scene datasets typically have obvious characteristics \cite{kitti, robotcar, msls}, such as the camera facing forward and the fixed viewing direction. Although these datasets may contain nighttime data, these characteristics do not suit VPR tasks in street scenes. Street scene datasets are usually obtained from the Google Street View download interface \cite{ gsv_cities, gsv, cosplace}, which can be arbitrarily expanded and includes almost all challenges of VPR tasks, except for nighttime scenes. Therefore, it is reasonable to conclude that the performance of existing VPR models in nighttime scenarios benefits from data augmentation during the training phase and the generalization capability during the inference phase.

\noindent \textbf{Nighttime Computer Vision} involves addressing classic downstream tasks using nighttime images, which can be categorized into two-stage and one-stage methods. Two-stage methods \cite{todaygan, gan_ss} involve converting nighttime images to daytime images before performing downstream tasks, while one-stage methodologies exploit raw nighttime data for training such tasks \cite{risp, vpr_bench}. For instance, Anoosheh et al. \cite{todaygan} propose a GAN method for converting nighttime images into daytime images and perform retrieval-based localization. Cui et al. \cite{risp} propose a method to perform reverse ISP and dark processing and train an end-to-end model for dark object detection. Xu et al. \cite{gan_ss} propose a GAN-based approach to convert nighttime images into daytime images for semantic segmentation of autonomous driving scenes. However, in the context of street-level VPR, two-stage methods suffer from poor generalization and increased computational requirements, whereas the one-stage approach is constrained by a scarcity of suitable training datasets. To address this limitation, we introduce a novel paradigm that integrates the two perspectives by transforming daytime images into nighttime images prior to VPR training.

\noindent \textbf{Image-to-Image Translation} is a class of vision and graphics problems where the goal is to learn the mapping between an input image and an output image using a training set. The classification of these methods is based on the type of training set, which can be pixel-level corresponding image-to-image or unpaired image-to-image \cite{unpaired, dualgan}. Obtaining large-scale, street-level, pixel-level correspondence day-night image pairs in the real world is an extremely challenging task. Therefore, we have considered the second type of training set. There are two main categories of methods used for unpaired image-to-image translation: two-side mapping and one-side mapping. The former, which includes CycleGAN \cite{unpaired} and DualGAN \cite{dualgan}, relies on the cycle-consistency constraint. This constraint ensures that the translated image can be reconstructed using an inverse mapping. However, the bijective projection required by this approach is often too limiting, which has led to the development of one-side mapping techniques. One such approach involves using a geometry distance to preserve content. DistanceGAN \cite{distance}, for instance, maintains the distances between images within domains, while GCGAN \cite{gcgan} ensures geometry-consistency between input and output. Another technique, CUT \cite{cut}, maximizes mutual information between the input and output using contrastive learning. In particular, the diffusion model \cite{egsde} has recently demonstrated outstanding performance in solving image-to-image translation problems. However, its usage necessitates substantial computational resources. After a thorough comparison of several approaches, we proceeded to train our day-to-night image-to-image translation network utilizing NEGCUT \cite{negcut} as the foundation.

\section{The Proposed Dataset}

In this chapter, we introduce a series of datasets that have been proposed, which are divided into two categories: NightStreet dataset for training a day-to-night image-to-image translation network and VPR-Night dataset for training night-to-day VPR networks.

\subsection{The NightStreet Dataset}

\begin{figure}[tbp] 
	\center
	{\includegraphics[width=0.43 \textwidth] {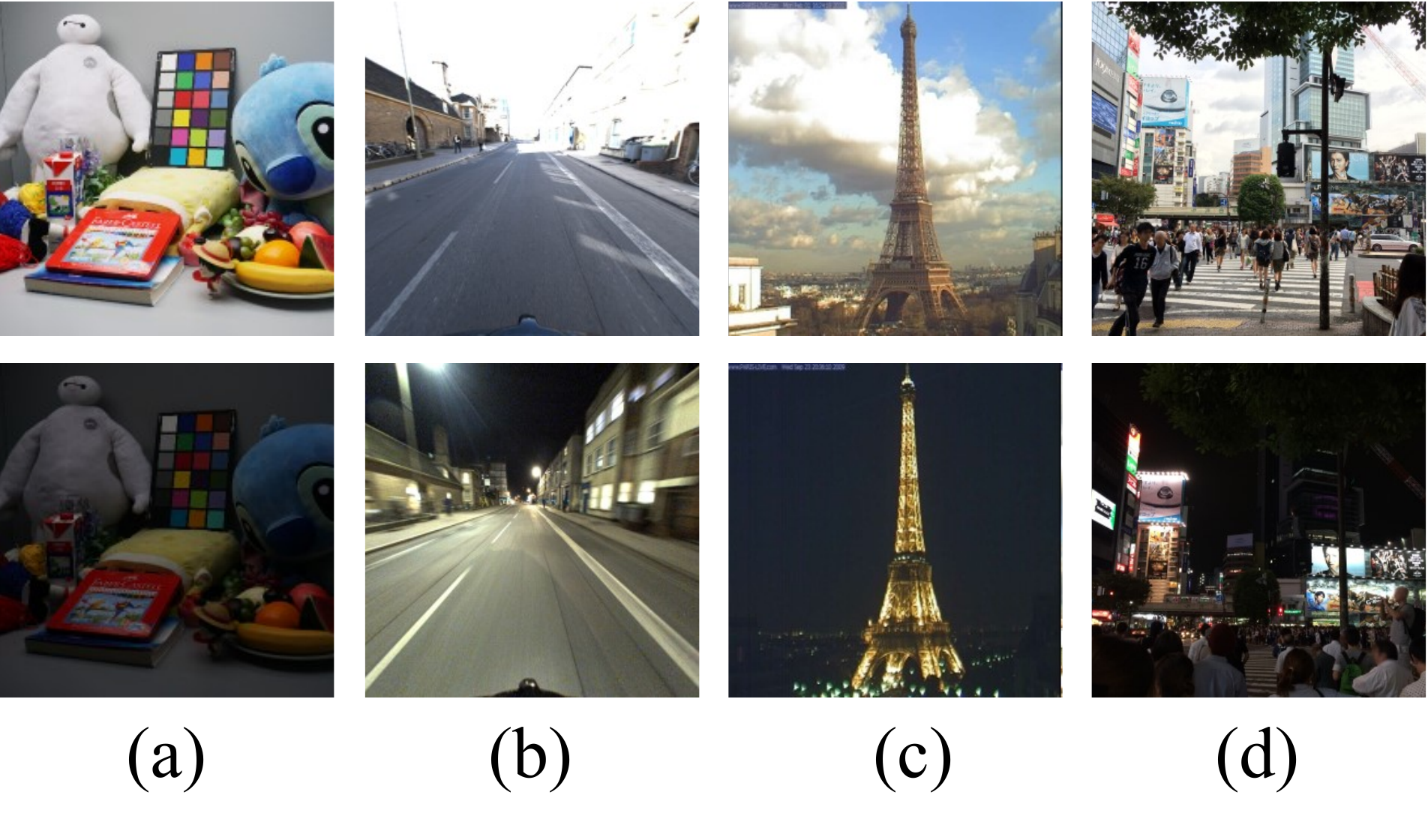}}
	\caption{ \textbf{Examples of Day-Night Style Datasets.} (a) is the LOL dataset \cite{lol_dataset}, (b) is the Robotcar dataset \cite{robotcar}, (c) is the Night2Day dataset \cite{night2day}, and (d) is the NightStreet dataset.} 
\label{dataset}
\end{figure}

Some datasets have been previously identified as potentially useful for day-to-night image-to-image translation. Nevertheless, they are not suitable for VPR tasks in street scenes. As depicted in Figure \ref{dataset} (a), the current low light enhancement research predominantly focuses on tackling low light conditions (such as those caused by backlighting) or weak exposure. However, utilizing these datasets in reverse fails to accurately capture the changes that occur during day-to-night translation. As demonstrated in Figure \ref{dataset} (b), there exist significant differences between automatic driving scenes and street scenes. For example, roadways typically constitute more than one-third of the image content, and images captured while the car is in motion tend to exhibit blurriness. Additionally, lighting conditions (e.g., street lights and car tail lights) generally exhibit limited variation. In regards to Figure \ref{dataset} (c), several time-lapse photography datasets have been proposed in the image generation field. However, these datasets share a common characteristic: the photographers who capture them tend to focus on distant views and skylines, which differ greatly from the urban street scenes.

We constructed day-night image pairs by directly rearranging the query images from Tokyo 24/7 and Aachen Day/Night datasets. Tokyo 24/7 \cite{247_dataset} provides 375 daytime and nighttime images each, while Aachen Day/Night \cite{vl_benchmarking} includes 234 daytime and 196 nighttime images. Importantly, we did not exploit the ground-truth relationship between the query and database images. Instead, we trained our translation network under unpaired images setting.

\subsection{The VPR-Night Datasets}

We used a trained day-to-night image-to-image translation model to process the existing VPR datasets and obtained the VPR-Night datasets. Theoretically, this method can be applied to any VPR dataset. In this study, we processed the Pitts-30k and SF-XL-small datasets, and named the new datasets Pitts-30k-N and SF-XL-small-N, respectively.

\noindent \textbf{Pitts-30k-N.} Pitts-30k is a subset of Pittsburgh-250k that is widely used in the research of VPR because of its high experimental efficiency \cite{netvlad}. It consists of 30k database images from Google Street View and 24k test queries generated from Street View but taken at different times, and is divided into three roughly equal parts for training, validation, and testing. According to the method designed in section \ref{pipelines}, we only need to perform day-to-night transfer on the test queries. Therefore, there are 24k night-style query images and 30k daytime database images in Pitts-30k-N dataset.

\noindent \textbf{SF-XL-small-N.} The San Francisco eXtra Large (SF-XL) \cite{cosplace} is a city-scale, dense, time-varying dataset. It crawls 3.43M equirectangular panoramas images from Google Street View and divides them into 12 crops, with the entire dataset consisting of 41.2M images. Each crop is labeled with 6 DoF information including GPS and heading orientation. Unlike the Pittsburgh dataset, the training set is not divided into query images and database images. To quickly validate our method, we opted to use a subset of SF-XL, namely SF-XL-small, comprising 100k street view images. Similarly, SF-XL-small-N contains 100k original images and 100k nighttime images.

\begin{figure*}[htbp] 
\begin{center}
{\includegraphics[width=0.7\textwidth] {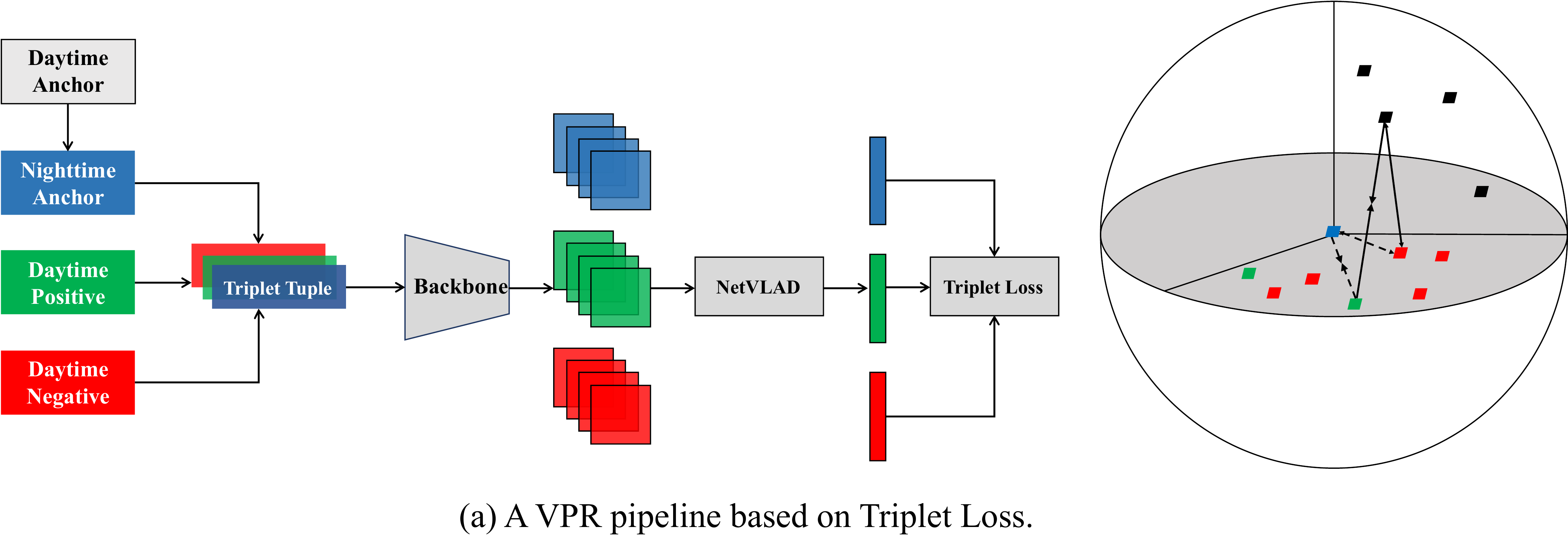}}
{\includegraphics[width=0.7\textwidth] {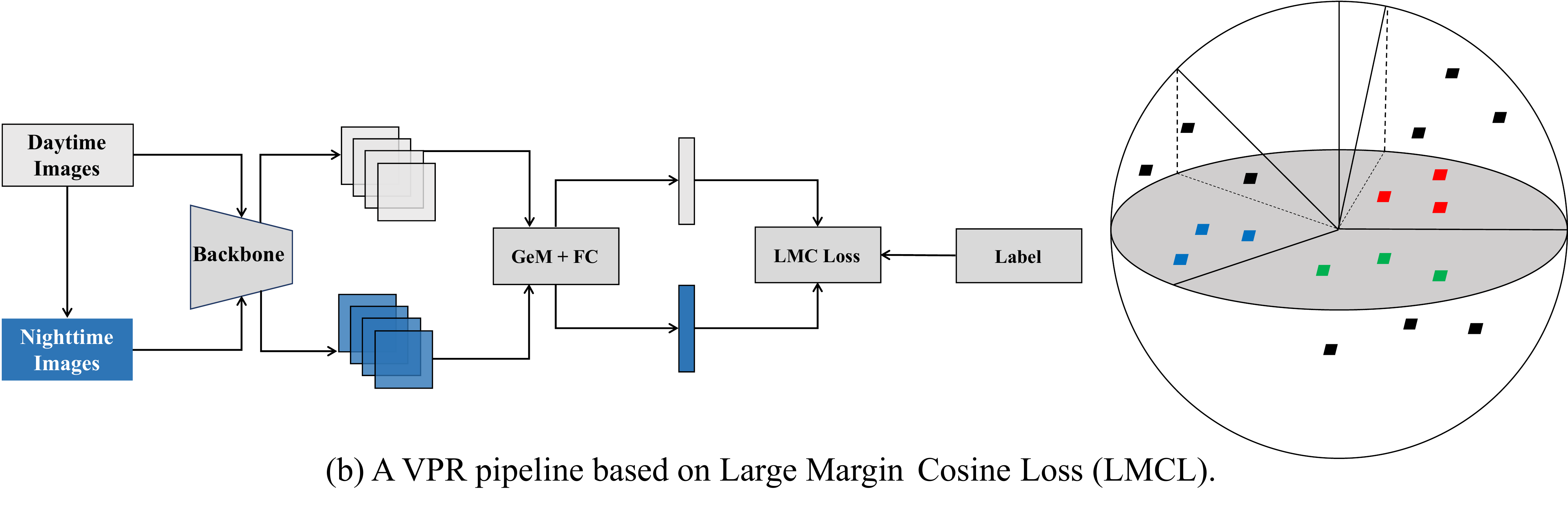}}
\end{center}
   \caption{\textbf{Schematic Diagrams of Two Popular VPR Pipelines.} In (a), the blue, green, and red blocks inside the sphere represent anchor, positive, and negative samples in the embedding space. In (b), the blue, green, and red blocks inside the sphere belong to different classes in the embedding space. In (a) and (b), the black blocks represent nighttime images.}
\label{fig:net}
\end{figure*}
\section{Method}

In this chapter, we first introduce the image-to-image translation from day-to-night, and then describe how to train two VPR pipelines on the generated nighttime data. Finally, we explain the rationale and implementation of dividing VPR and conquering NPR.

\subsection{Day-to-Night Image-to-Image Translation}

In this section, we introduce a contrastive learning-based unpaired image-to-image translation network \cite{cut, negcut}, which is trained on the NightStreet dataset and generates the VPR-Night datasets. Our goal is to preserve the content of daytime images while specifying nighttime style. We define the daytime images from the input domain as $\mathcal{X} \subset \mathbb{R}^{H \times W \times 3}$ and the nighttime images from the output domain as $\mathcal{Y} \subset \mathbb{R}^{H \times W \times 3}$, and aim to learn a mapping $G: \mathcal{X} \to \mathcal{Y}$. The NightStreet dataset can be represented as an instance $X = \{x \in \mathcal{X}\}$ and $Y = \{y \in \mathcal{Y}\}$. The mapping function $G$ is decomposed into an encoder $G_{\mathrm{enc}}$ and a generator $G_{\mathrm{dec}}$, so the process of producing output images can be represented as: 
\begin{equation}
\hat{y} = G(x) = G_{\mathrm{dec}}(G_{\mathrm{enc}}(x)).
\end{equation}
We encourage the output images to have a nighttime style similar to the target domain by using a discriminator $D$ and the following adversarial loss \cite{unpaired}: 
\begin{equation}
\mathcal{L}_{\mathrm{GAN}} = \mathbb{E}_{y\sim Y}\log{D(y)} + \mathbb{E}_{x\sim X}\log{(1-D(G(x)))}.
\end{equation}
Then, a contrastive learning framework is employed to maintain local content consistency between input $x$ and output $\hat{y}$. We extract a certain number $N$ of patches from the images, where one pair of patches is located at the same position in $x$ and $\hat{y}$, denoted as $q$ and $p$, while the remaining $N-2$ patches are randomly selected from the $x$. These patches are then fed into a feature extraction network to obtain feature vectors, and an $(N-1)$ classification problem is established. The feature extraction network used here is based on the $G_{\mathrm{enc}}$, followed by a 2-layer MLP.

\subsection{Two Night-to-Day VPR Pipelines}
\label{pipelines}

In this section, we introduce strategies for utilizing nighttime data to train two popular VPR frameworks, namely the triplet network shown in Figure \ref{fig:net} (a) and the classification network shown in Figure \ref{fig:net} (b).

\noindent \textbf{Triplet Network.} From the training set with GPS, anchor samples $q$ are selected, and data sharing the same GPS label or with close proximity are considered positive samples $p$, while the remaining samples are regarded as negative samples $n$. These samples are then fed into $f$, a pre-trained deep neural network with an aggregation layer, obtaining feature vectors $f(q), f(p),$ and $f(n)$ in the embedding space. Finally, the Euclidean distance between the $f(q)$ and $f(p)$, as well as $f(q)$ and $f(n)$, are calculated in the embedding space to obtain a triplet loss, which is formulated as:
\begin{equation}
L_{\mathrm{triplet}} = l(||f(q)-f(p)||_2^2 - ||f(q)-f(n)||_2^2 + m),
\end{equation}
where $l$ is the hingle loss $l(x) = \mathrm{max}(x, 0)$, $m$ is the margin parameter that controls the distance between samples of the same class and different classes.

Considering that we aim to achieve Night-to-Day VPR, we need to construct sample pairs with different styles, i.e., we need to transfer the anchor samples to nighttime style or convert both positive and negative samples to nighttime style. The latter method is obviously less efficient than the former.

\noindent \textbf{Classification Network.} VPR can be treated as a classification problem based on labels. Following \cite{cosplace}, the training set can be partitioned into classes by splitting it into square geographical cells using UTM coordinates $\{\textit{east}, \textit{north}\}$ and further dividing each cell into a set of classes based on the orientation$\{\textit{heading}\}$ of each image. We transformed all images in the database into a night style while preserving their original labels. Then we employed the Large Margin Cosine Loss (LMCL) \cite{cosface} to train a model:
\begin{equation}
 L_{\mathrm{lmc}} = \frac{1}{N}\sum_{i}{-\log{\frac{e^{s (\cos(\theta_{{y_i}, i}) - m)}}{e^{s (\cos(\theta_{{y_i}, i}) - m)} + \sum_{j \neq y_i}{e^{s \cos(\theta_{j, i})}}}}},
\end{equation}
subject to
\begin{equation}
\begin{split}
  cos(\theta_j,i) &= {W_j}^Tx_i,
\end{split}
\end{equation}
where $N$ is the numer of training images, $x_i$ is the $i$-th embedding vector corresponding to the ground truth class of $y_i$, the $W_j$ is the weight vector of the $j$-th class, and $\theta_j$ is the angle between $W_j$ and $x_i$. $s$ and $m$ are two hyperparameters that control the weights of the intra-class distance and inter-class distance in the loss function, respectively.

\subsection{Dividing VPR and Conquering NPR}

In this section, we introduce concepts from three fields: deep learning, neuroscience, and computer algorithms. Our aim is to explain the rationale behind the need to divide night vision problem from general visual problem\footnote{Although we focus solely on the VPR task, we believe that this framework should be applicable to many other computer vision tasks.}. 

\noindent \textbf{Deep learning.} We utilize data-driven network learning, where the training set and test set should have similar distributions \cite{deeplearning}. However, previous research on nighttime VPR does not follow this principle, which is the reason why all methods degrade severely in nighttime scenes. When the fitting ability of the model is limited, increasing the amount of nighttime data will also cause the VPR performance in daytime scenes to degrade.

\noindent \textbf{Neuroscience.} There are two types of photoreceptor cells distributed on the retina: cone cells and rod cells \cite{neural_vision}. Cone cells are primarily responsible for color and detail discrimination and are only activated under relatively adequate lighting conditions. Rod cells, on the other hand, are mainly responsible for identifying the intensity of light and motion, and can be responsive under low-light conditions. Nocturnal animals possess a greater number of rod cells.

\noindent \textbf{Computer algorithm.} Divide-and-conquer \cite{intro_algorithm} (D\&C) is a classic algorithm that decomposes a large problem into several smaller but structurally similar sub-problems, recursively solving these sub-problems, and then combining the sub-problem solutions, to obtain the original problem solution. 

Based on the above knowledge, it is suggested that the original model should be used for daytime scenes, while the VPR-Night trained or fine-tuned model should be used for nighttime scenes. Then we provide three implementation ideas for the "divide" step: i) training a day-night network for classification, ii) using the global average brightness and an empirical threshold for classification, and iii) using system time and local sunset time \footnote{\href{https://www.timeanddate.com/astronomy/}{https://www.timeanddate.com/astronomy/}} for classification. The first method can achieve better results, but has low practicality in real-world applications. The second method may fail in some specific situations, such as scenes with strong backlight being misclassified as night and scenes with bright light being misclassified as day. The third method is independent of image content, but is only applicable to devices equipped with communication functions, such as smartphones and robots. In our experiments, we use a combination of the second and third methods for decomposing the test set. Finally, we emphasize that the execution of "divide" in the real world is a one-time process, so the additional computation required for new pipelines can be ignored.

\section{Experiments}

\begin{table*}[ht]
\centering
\resizebox{\textwidth}{!}{
   \begin{tabular}{l l l l l l l l l l}
    \toprule
    \multirow{2}{*}{Method} &
    \multirow{2}{*}{Backbone} & 
    \begin{tabular}[c]{@{}l@{}}Aggregation \\ Method           \end{tabular} & 
    \begin{tabular}[c]{@{}l@{}}Feature     \\ Dim              \end{tabular} &
    \begin{tabular}[c]{@{}l@{}}Loss        \\ Function         \end{tabular} &
    \begin{tabular}[c]{@{}l@{}}Training    \\ Dataset          \end{tabular} &
    \begin{tabular}[c]{@{}l@{}}R@1         \\ All queries      \end{tabular} & 
    \begin{tabular}[c]{@{}l@{}}R@1         \\ Day queries      \end{tabular} & 
    \begin{tabular}[c]{@{}l@{}}R@1         \\ Sunset queries   \end{tabular} & 
    \begin{tabular}[c]{@{}l@{}}R@1         \\ Night queries    \end{tabular} \\
    \midrule
    \begin{tabular}[c]{@{}l@{}l@{}} NetVLAD \cite{netvlad} (\textit{CVPR'16}) \\ NetVLAD-NPR  \\ NetVLAD-D\&C \end{tabular} &
    \begin{tabular}[c]{@{}l@{}l@{}} &                                         \\ VGG-16       \\ &            \end{tabular} &
    \begin{tabular}[c]{@{}l@{}l@{}} &                                         \\ NetVLAD      \\ &            \end{tabular} &
    \begin{tabular}[c]{@{}l@{}l@{}} &                                         \\ 32768        \\ &            \end{tabular} &
    \begin{tabular}[c]{@{}l@{}l@{}} &                                         \\ Triplet Loss \\ &            \end{tabular} &
    \begin{tabular}[c]{@{}l@{}l@{}} Pitts-30k                                 \\ Pitts-30k-N  \\ -            \end{tabular} & 
    \begin{tabular}[c]{@{}l@{}l@{}} 63.8 \\ 62.5 (\textcolor{blue}{-1.3})  \\ 68.6 (\textcolor{red}{+4.8})    \end{tabular} & 
    \begin{tabular}[c]{@{}l@{}l@{}} 88.6 \\ 80.0 (\textcolor{blue}{-8.6})  \\ 88.6                            \end{tabular} &
    \begin{tabular}[c]{@{}l@{}l@{}} 75.2 \\ 64.8 (\textcolor{blue}{-10.4}) \\ 74.3 (\textcolor{blue}{-0.9})   \end{tabular} & 
    \begin{tabular}[c]{@{}l@{}l@{}} 27.6 \\ 42.9 (\textcolor{red}{+15.3})  \\ 42.9 (\textcolor{red}{+15.3})   \end{tabular} \\
    \midrule
    NetVLAD \cite{netvlad} (\textit{CVPR'16}) & 
    VGG-16 & 
    NetVLAD+PCA & 
    4096 &
    Triplet Loss &
    Pitts-30k & 
    68.9 & 
    - &
    - & 
    - \\
    \midrule
    DIR \cite{dir} (\textit{T-PAMI'18}) &
    Res-101 & 
    GeM+FC & 
    2048 &
    Triplet Loss &
    Google Landmark & 
    74.9 & 
    92.4 &
    81.9 & 
    50.5 \\
    \midrule
    SARE \cite{sare} (\textit{ICCV'19}) &
    VGG-16 & 
    NetVLAD+PCA & 
    4096 &
    SARE-Joint &
    Pitts-30k & 
    74.8 & 
    -  &
    -  & 
    -  \\
    \midrule
    SFRS \cite{sfrs} (\textit{ECCV'20}) &
    VGG-16 & 
    NetVLAD+PCA & 
    4096 &
    SARE-Joint &
    Pitts-30k  & 
    78.5 & 
    -    &
    -    & 
    - \\
    \midrule
    APPSVR \cite{appsvr} (\textit{ICCV'21}) &
    VGG-16 & 
    APP+PCA & 
    4096 &
    Triplet Loss &
    Pitts-30k & 
    77.1 & 
    - &
    - & 
    - \\
    \midrule
    \begin{tabular}[c]{@{}l@{}l@{}}DVG \cite{dvg} (\textit{CVPR'22})\\ DVG-NPR \\ DVG-D\&C \end{tabular} &
    \begin{tabular}[c]{@{}l@{}l@{}} & \\ Res-18 \\ & \end{tabular} &
    \begin{tabular}[c]{@{}l@{}l@{}} & \\ NetVLAD \\ & \end{tabular} & 
    \begin{tabular}[c]{@{}l@{}l@{}} & \\ 16384 \\ & \end{tabular} & 
    \begin{tabular}[c]{@{}l@{}l@{}} & \\ Triplet Loss \\ & \end{tabular} & 
    \begin{tabular}[c]{@{}l@{}l@{}}Pitts-30k\\ Pitts-30k-N \\ - \end{tabular} & 
    \begin{tabular}[c]{@{}l@{}l@{}}66.7 \\ 74.0 (\textcolor{red}{+7.3})\\ 74.9 (\textcolor{red}{+8.2}) \end{tabular} & 
    \begin{tabular}[c]{@{}l@{}l@{}}90.5 \\ 85.7 (\textcolor{blue}{-4.8})\\ 90.5   \end{tabular} &
    \begin{tabular}[c]{@{}l@{}l@{}}76.2 \\ 80.0 (\textcolor{red}{+3.8}) \\ 78.1 (\textcolor{red}{+1.9}) \end{tabular} & 
    \begin{tabular}[c]{@{}l@{}l@{}}33.3 \\ 56.2 (\textcolor{red}{+22.9}) \\ 56.2 (\textcolor{red}{+22.9})  \end{tabular} \\
    \midrule
    \begin{tabular}[c]{@{}l@{}l@{}} CosPlace \cite{cosplace} (\textit{CVPR'22}) \\ CosPlace-NPR \\ CosPlace-D\&C \end{tabular} &
    \begin{tabular}[c]{@{}l@{}l@{}} &                                           \\ VGG-16                   \\ & \end{tabular} &
    \begin{tabular}[c]{@{}l@{}l@{}} &                                           \\ GeM+FC                   \\ & \end{tabular} &
    \begin{tabular}[c]{@{}l@{}l@{}} &                                           \\ 512                      \\ & \end{tabular} &
    \begin{tabular}[c]{@{}l@{}l@{}} &                                           \\ LCM Loss                 \\ & \end{tabular} &
    \begin{tabular}[c]{@{}l@{}l@{}} SF-XL \\ Fine tuning on small-N \\ - \end{tabular} & 
    \begin{tabular}[c]{@{}l@{}l@{}}  81.9 \\ 82.9 (\textcolor{red}{+1.0})  \\  84.1 (\textcolor{red}{+2.2}) \end{tabular} & 
    \begin{tabular}[c]{@{}l@{}l@{}}  90.5 \\ 87.6 (\textcolor{blue}{-2.9}) \\ 90.5 \end{tabular} &
    \begin{tabular}[c]{@{}l@{}l@{}}  89.5 \\ 83.8 (\textcolor{blue}{-5.7}) \\ 84.8 (\textcolor{blue}{-4.7})\end{tabular} & 
    \begin{tabular}[c]{@{}l@{}l@{}}  65.7 \\ 77.1 (\textcolor{red}{+11.4}) \\ 77.1 (\textcolor{red}{+11.4}) \end{tabular} \\
    \midrule
    \begin{tabular}[c]{@{}l@{}l@{}} CosPlace \cite{cosplace}  \\ CosPlace-NPR            \end{tabular} &
    \begin{tabular}[c]{@{}l@{}l@{}} &                         \\ Res-50                  \end{tabular} &
    \begin{tabular}[c]{@{}l@{}l@{}} &                         \\ GeM+FC                  \end{tabular} &
    \begin{tabular}[c]{@{}l@{}l@{}} &                         \\ 512                     \end{tabular} &
    \begin{tabular}[c]{@{}l@{}l@{}} &                         \\ LCM Loss                \end{tabular} &
    \begin{tabular}[c]{@{}l@{}l@{}} SF-XL \\ Fine tuning on small-N       \end{tabular} & 
    \begin{tabular}[c]{@{}l@{}l@{}}  88.6 \\ 92.1 (\textcolor{red}{+3.5}) \end{tabular} & 
    \begin{tabular}[c]{@{}l@{}l@{}}  95.2 \\ 96.2 (\textcolor{red}{+1.0}) \end{tabular} &
    \begin{tabular}[c]{@{}l@{}l@{}}  90.5 \\ 93.3 (\textcolor{red}{+2.8}) \end{tabular} & 
    \begin{tabular}[c]{@{}l@{}l@{}}  80.0 \\ 86.7 (\textcolor{red}{+6.7}) \end{tabular} \\
    \bottomrule
    \\
  \end{tabular}}\\

\caption{\textbf{Comparisons of various methods on Tokyo 24/7 \cite{247_dataset}.} We reproduced three methods and added their NPR and D\&C versions. As previously mentioned, none of the previous methods conducted experiments on the last three columns. Therefore, we currently only report their Recall@1 metrics for all queries.}
\label{tab:vpr}
\end{table*}

In this chapter, we describe the implementation details, the test set, and the evaluation methods. We provide quantitative and qualitative results, followed by visualizations that demonstrate the utility of NPR as well as its current limitations.

\subsection{Implementation details}

\noindent \textbf{Day-to-Night Image-to-Image Translation.} We use the same training parameters as in \cite{negcut} and select the model from the 400th epoch for inference. Since the image size of NightStreet is complex and differs greatly from that of the VPR training set, we scale the NightStreet images proportionally with the shorter side fixed at 640 and then randomly crop them to a size of 512$\times$512 during training. This can maintain scale consistency between the training set and testing set as much as possible. The translation network takes approximately one day to process the SF-XL-small dataset, at a resolution of 512$\times$512 on a single 3090 Ti.

\noindent \textbf{Night-to-Day Visual Place Recognition.} The essence of our method is to transfer the nighttime style to existing datasets and combine it with existing pipelines. Therefore, it can be compatible with any VPR method and improve their performance in nighttime scenes. We replicated the work of \cite{netvlad, dvg, cosplace}, and applied our method. Since most of the previous methods used the VGG-16 structure, we trained and compared it with this structure. The current top-ranked method on Tokyo 24/7 is based on ResNet-50, so we also conducted experiments based on it. We reproduce the best performance of the baseline method or directly use the model open-sourced by the author. Specifically, data augmentation was not used during NPR training and the learning rate used when fine-tuning the model was 1e-6.

\noindent \textbf{Night-to-Day Visual Localization.} We adopted the hierarchical localization framework provided by \cite{hloc} and replaced the VPR module with Superpoint \cite{superpoint} and Superglue \cite{superglue} for local feature extraction and matching, respectively. Notably, we adjusted the input image resolution of the VPR module to the recommended parameters for this project.

\subsection{Datasets and evaluation methodology}

We reported results on multiple publicly available datasets.

\noindent \textbf{Tokyo 24/7 v2} \cite{247_dataset} contains 75,984 database images from Google Street View and 315 query images taken using mobile phone cameras. This is an extremely challenging dataset where the queries were taken at daytime, sunset, nighttime, while the database images were only taken at daytime. However, to the best of our knowledge, all works using this dataset have not tested them by category. Even the original work that proposed this dataset evaluated sunset and night as one category. This incident seriously overshadowed the fact that the performance of VPR was poor under the cover of night. We propose two ways to test the dataset: one is to directly split by labels, and the other is to use the information of exchangeable image file format (EXIF) and local sunset time \footnote{\href{https://www.timeanddate.com/sun/japan/tokyo?month=9&year=2014}{https://www.timeanddate.com/sun/japan/tokyo?month=9\&year=2014}} for partitioning. The second method will split the sunset testing set into two categories, test them separately, and then merge the results.

\noindent \textbf{Aachen Day/Night v1} \cite{vl_benchmarking} comprises 922 query images captured during day and night, and 4328 reference images collected over a span of two years using hand-held cameras. All query images were captured using mobile phone cameras, hence the Aachen Day-Night dataset considers the scenario of localization using mobile devices, e.g., for augmented or mixed reality applications.

\noindent \textbf{Other datasets.} Additionally, we employ the Pitts-30k and SF-XL-test-v1 datasets to investigate the phenomenon of model degradation.

\noindent \textbf{Evaluation metric.} In the VPR dataset, we use the recall@N with a 25 meters threshold, i.e., the percentage of queries for which at least one of the first N predictions is within a 25 meters distance from the ground truth position of the query, following standard procedure \cite{netvlad, 247_dataset, appsvr, cosplace, sare, sfrs}. In the visual localization dataset, we evaluate VPR using the success rate of localization under different recall@N metrics. Specifically, the visual localization system considers higher localization precision than that required by the VPR dataset.

\subsection{Comparison with the State-of-the-art Methods}

\begin{figure}
\begin{center}
{\includegraphics[width=0.235\textwidth] {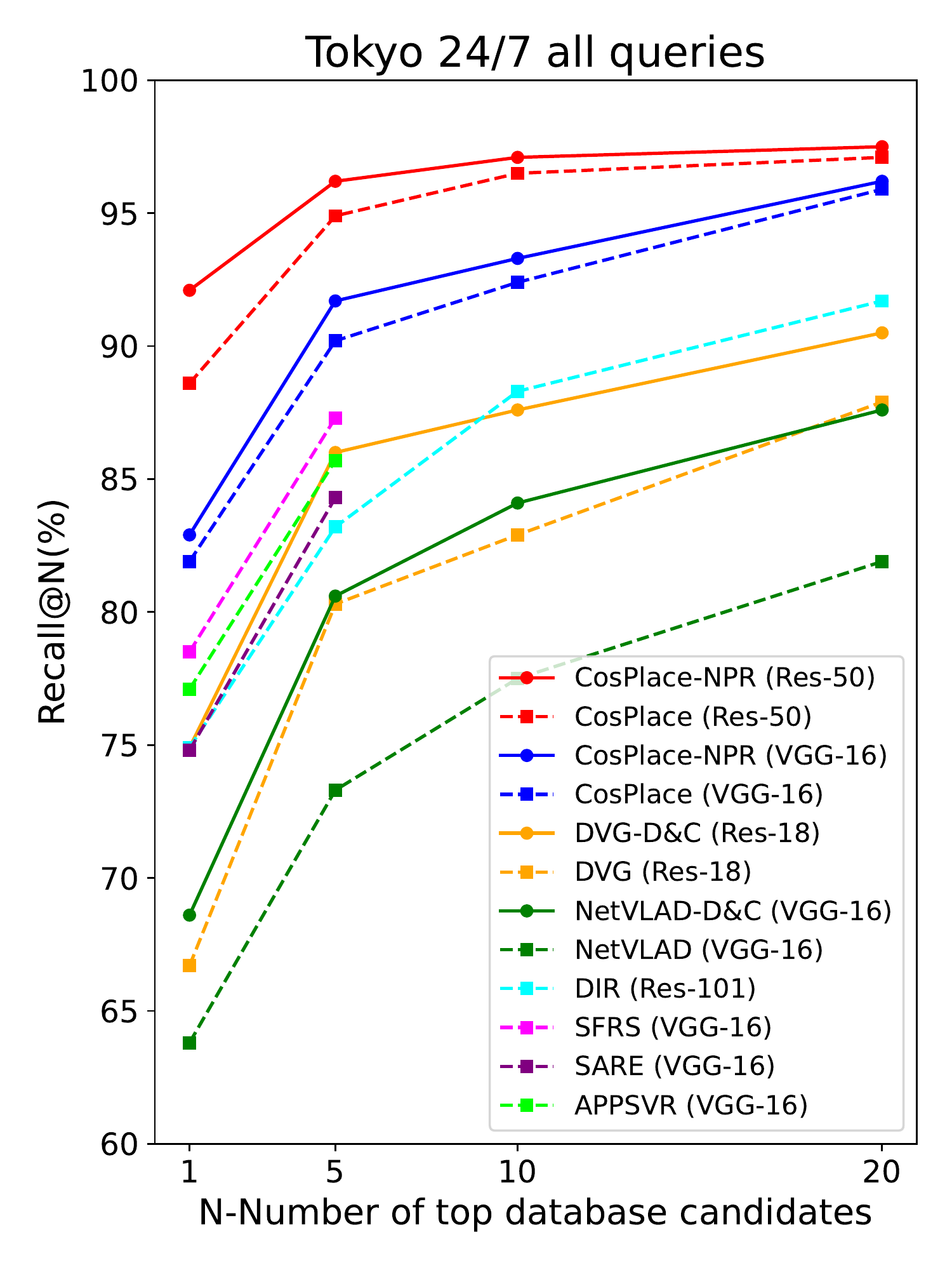}
\includegraphics[width=0.235\textwidth] {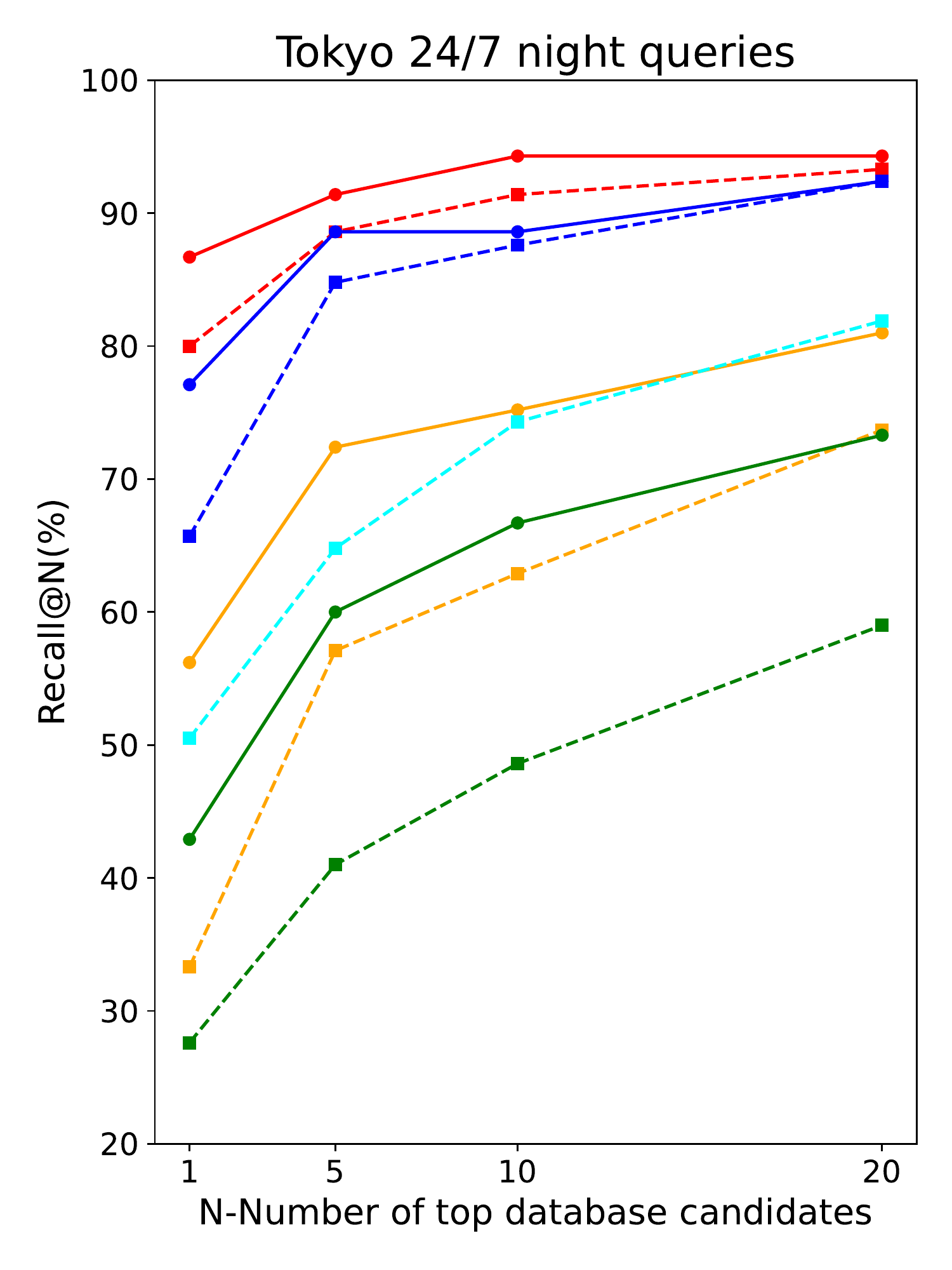}}
{}
\end{center}
   \caption{\textbf{Comparison of our methods and baseline.} The solid line represents our method, while the dashed line represents the baseline method. It should be noted that the y-axis origins of the two graphs are different.}
\label{fig:R@N}
\end{figure}

\begin{table*}[ht]
  \centering
  \resizebox{\textwidth}{!}{
   \begin{tabular}{llllllllll}
    \toprule
    \multirow{2}{*}{Method} &
    \multirow{2}{*}{Backbone} & 
    \multirow{2}{*}{\begin{tabular}[c]{@{}l@{}}Aggregation\\ Method\end{tabular}} &
    \multirow{2}{*}{\begin{tabular}[c]{@{}l@{}}Feature \\ Dim\end{tabular}} & 
    \multicolumn{4}{c}{(0.25m, 2\degree)   /   (0.5m, 5$\degree$)   /   (5m, 10\degree)} \\
    & & & &  R@1 & R@5 & R@10 & R@20
    \\
    \midrule
    \begin{tabular}[c]{@{}l@{}l@{}} NetVLAD \cite{netvlad} \end{tabular} &
    VGG-16 & 
    NetVLAD+PCA & 
    4096 &
    \begin{tabular}[c]{@{}l@{}l@{}} 68.4 / 78.6 / 85.7 \\   \end{tabular} & 
    \begin{tabular}[c]{@{}l@{}l@{}} 75.5 / 84.7 / 93.9 \\   \end{tabular} &
    \begin{tabular}[c]{@{}l@{}l@{}} \textcolor{red}{82.7} / 88.8 / 95.9\\   \end{tabular} & 
    \begin{tabular}[c]{@{}l@{}l@{}} 83.7 / \textcolor{red}{92.9} /  \textcolor{red}{99.0} \\   \end{tabular} \\
    \midrule
    \begin{tabular}[c]{@{}l@{}l@{}}CosPlace \\ CosPlace-NPR \end{tabular} &
    Res-50 & 
    GeM &
    512 &
    \begin{tabular}[c]{@{}l@{}l@{}} 69.4 / 82.7 / 89.8 \\  \textcolor{red}{71.4} / \textcolor{red}{84.7} / \textcolor{red}{93.9} \end{tabular} & 
    \begin{tabular}[c]{@{}l@{}l@{}} 77.6 / 87.8 / 95.9 \\  \textcolor{red}{78.6} / \textcolor{red}{88.8} / 95.9 \end{tabular} &
    \begin{tabular}[c]{@{}l@{}l@{}} 77.6 / 87.8 / 95.9 \\ 79.6 / 88.8 / 95.9 \end{tabular} & 
    \begin{tabular}[c]{@{}l@{}}  83.7 / 90.8 / 98.0    \\  \textcolor{red}{84.7} / 91.8 / 98.0  \end{tabular} \\
    \bottomrule
    \\
  \end{tabular}}
    \caption{\textbf{Comparisons of three methods on Aachen Day/Night \cite{long_vl}.} All three methods employ the same VL pipeline \cite{hloc}, with only the VPR method being replaced. The success rates under different levels of localization accuracy were evaluated at Recall@N.}
\label{vl}
\end{table*}

As shown in Table \ref{tab:vpr}, we have replicated the experimental results of NetVLAD \cite{netvlad}, DIR \cite{dir}, DVG \cite{dvg}, and CosPlace \cite{cosplace} on the Tokyo 24/7 dataset, while citing the results of NetVLAD with PCA \cite{netvlad}, SARE \cite{sare}, SFRS \cite{sfrs}, and APPSVR \cite{appsvr} from the Deep Visual Geo-Localization Benchmark \cite{dvg}. Our method was applied to these three replication methods and labeled with the name suffixes NPR and D\&C accordingly. The results can be summarized in a few points:

\noindent 1) The Recall@1 for nighttime queries are significantly lower than those for daytime queries across all methods. The Recall@1 for all queries is a trade-off result. This confirms our viewpoint that the challenge of Nighttime VPR has been overlooked for a considerable period of time.

\noindent 2) All methods trained on the VPR-Night datasets showed a significant improvement in performance on nighttime queries. Models with weaker fitting abilities, such as VGG-16 and ResNet-18, showed a corresponding degradation in daytime scenes, whereas networks with stronger fitting abilities, such as ResNet-50, were able to improve Recall@1 for both daytime and nighttime queries.

\noindent 3) To address the issue of imbalanced performance between daytime and nighttime for small models, a divide-and-conquer algorithm can effectively maintain performance balance, which is particularly beneficial for models deployed on mobile platforms.

As illustrated in Figure \ref{fig:R@N}, we present the variation of accuracy with respect to different recall values, N. Our proposed method shows a significant performance improvement over the baseline approach across all recall values. It is worth noting that the vertical axes of the two plots have different origins.

As shown in Table \ref{vl}, our proposed method outperforms the baseline approach on the nighttime testing subset of the Aachen dataset. While our localization success rates at R@10 and R@20 are slightly lower than those of NetVLAD with PCA, we think this may be attributed to the fact that the size of the database cannot fully reflect the advantages of our model, and our output dimensionality is also significantly lower than that of NetVLAD.

\subsection{Daytime VPR experiments}

\begin{table}
  \centering
  \resizebox{0.48\textwidth}{!}{
\begin{tabular}{lllll}
\toprule
             & \multicolumn{2}{l}{Pitts-30k-test} & \multicolumn{2}{l}{SF-XL-test-v1}  \\
             & R@1  & R@5                    & R@1  & R@5                         \\
\midrule
CosPlace    & 88.4 & 94.6                   & 68.1 & 78.9                        \\
CosPlace-NPR & 88.0 (\textcolor{blue}{-0.4})  & 94.1 (\textcolor{blue}{-0.5}) &
                65.4 (\textcolor{blue}{-2.7})& 75.8 (\textcolor{blue}{-3.1})             \\
\bottomrule\\
\end{tabular}}
\caption{\textbf{Results on Pitts-30k-test \cite{netvlad} and SF-XL-test-v1 \cite{cosplace}.} Our Model is CosPlace+NPR (VGG-16), while the comparative method is CosPlace (VGG-16). It should be noted that for testing SF-XL-test-v1, we used the SF-XL-small database instead of SF-XL. Therefore, the CosPlace's result presented here are slightly better than that reported in \cite{cosplace}.}
\label{daytime}
\end{table}

While we did not have high expectations for the performance of the NPR during daytime, we still demonstrated its performance on other daytime datasets and compared it with baseline methods. As shown in Table \ref{daytime}, our method exhibits a slight decrease in performance on the Pitts-30k and SF-XL-test-v1 datasets.

\subsection{Qualitative experiment}

\begin{figure*}
\begin{center}
{\includegraphics[width=0.85\textwidth] {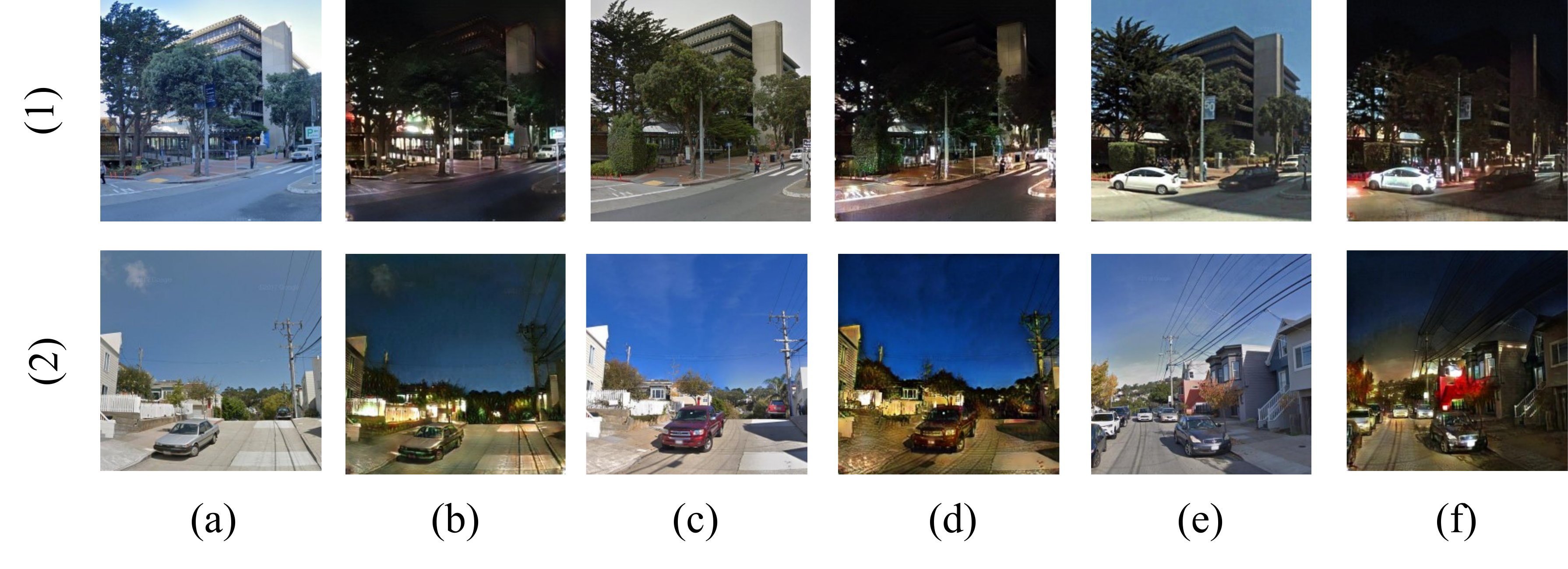}}
\end{center}
   \caption{\textbf{Examples of translation results from SF-XL-small-N.} Each row corresponds to one location. (a), (c), and (e) represent the images captured at different times for the same location, while (b), (d), and (f) correspond to the nighttime images generated from (a), (c), and (e), respectively. }
\label{fig:gan}
\end{figure*}

\begin{figure*}[h]
\begin{center}
{\includegraphics[width=0.85\textwidth] {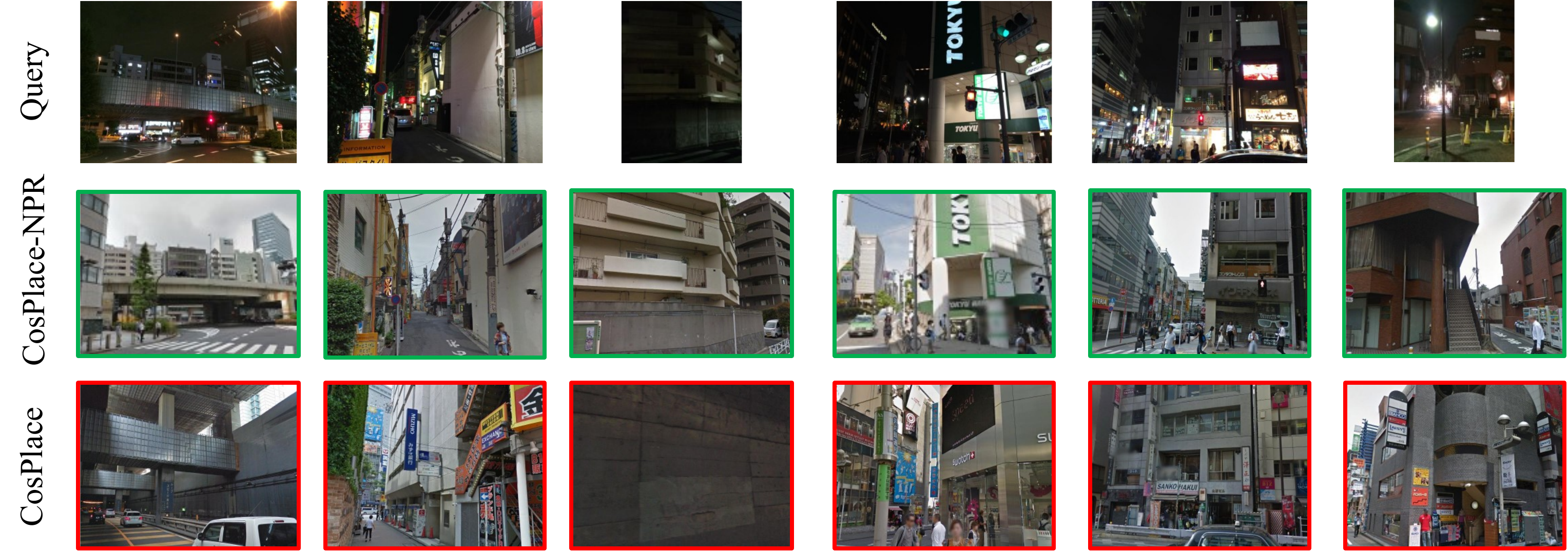}}
\end{center}
   \caption{\textbf{Examples of retrieval results for challenging queries on Tokyo 24/7.} Each column corresponds to one query case: the query is shown in the first row, the top retrieved image using our best method (CosPlace-NPR) in the second, and the top retrieved image using our best baseline (CosPlace) in the last row. Correct retrievals are indicated with a green border, while incorrect retrievals are indicated with a red border.}
\label{fig:demo2}
\end{figure*}

In this section, we present samples from the generated VPR-Night datasets and the results of CosPlace-NPR. As shown in Figure \ref{fig:gan}, (1) and (2) represent two locations, (a), (c), and (e) are data from different times of day in Google Street View, while (b), (d), and (f) are generated results. Our method preserves scale, appearance, and viewpoint changes while adding a nighttime style.

As shown in Figure \ref{fig:demo2}, our incremental improvement significantly enhances performance at night. These nocturnal query images not only exhibit extreme illumination differences with database images, but also undergo architectural resets, changes in perspective, and scale variations. This is highly consistent with real-world scenarios.

\subsection{Limitations}

We believe that our theory is not limited, but the current implementation is constrained. i) We require a larger NightStreet dataset to capture a richer range of day-to-night variations. Fortunately, the loose requirement for unpaired images makes the dataset easily extensible. And constructing the NightStreet dataset is undoubtedly less challenging than creating a large-scale, street-level, day-to-night corresponding VPR dataset. ii) Rendering large-scale datasets such as SF-XL requires significant GPU resources. We plan to address both of these limitations in our future work.

\section{Conclusions}

In this work, we address the challenging problem of nighttime VPR, which has been hindered by the lack of appropriate training datasets and inaccurate testing methodologies. To overcome these issues, we propose a dedicated pipeline for Noctural Place Recognition. First, we construct the NightStreet dataset and train a day-to-night image-to-image translation network. We then apply the network to process existing large-scale VPR datasets and demonstrate how to integrate them into two popular VPR pipelines. Finally, we introduce the idea of differentiating between VPR and NPR, providing a multidimensional interpretation. Our experimental results show that our pipeline significantly improves previous methods.

{\small
\bibliographystyle{ieee_fullname}
\bibliography{egbib_for_review}
}

\end{document}